\documentclass[11pt,a4paper]{article}
\usepackage[hyperref]{emnlp-ijcnlp-2019}
\usepackage{times}
\usepackage{latexsym}
\usepackage{amssymb}
\usepackage{url}
\usepackage{tikz}
\usepackage{cjhebrew} 
\usepackage{graphicx} 

\aclfinalcopy

\title{What's Wrong with Hebrew NLP? \\ And How to Make it Right}

\author{Reut Tsarfaty\hspace{0.2in}Amit Seker\hspace{0.2in}Shoval Sadde\hspace{0.2in}Stav Klein \\
  Open University of Israel, University Road 1, Ra'anana, Israel \\
  {\tt \{reutts,shovalsa,amitse,stavkl\}@openu.ac.il} 
  }

\date{}

\begin{document}
\maketitle
 
\begin{abstract}

For languages with simple morphology, such as English, automatic annotation pipelines such as spaCy or Stanford's CoreNLP successfully serve projects in academia and the industry. 
For many morphologically-rich languages (MRLs), similar pipelines show sub-optimal performance that limits their applicability for text analysis in research and the industry.
The sub-optimal performance is mainly due to errors in early morphological disambiguation decisions, which cannot be recovered later in the pipeline, yielding incoherent  annotations on the whole.  In this paper we describe the design and use of the {\sc Onlp} suite,  a joint morpho-syntactic parsing framework for processing Modern Hebrew texts. The  joint inference over morphology and syntax substantially limits error propagation, and leads to high accuracy.  {\sc Onlp}  provides rich and expressive output which already serves diverse academic  and commercial needs. Its accompanying online demo further serves educational activities,  introducing Hebrew NLP intricacies to researchers and non-researchers alike.

\end{abstract}

\section{Introduction}
\label{sec:Intro}

NLP pipelines for the automatic annotation of unstructured texts are at the core of language technology applications  for Data Science, Text Analytic and Artificial Intelligence. 
For English,  annotation pipelines such as spaCy \cite{spacy} or Stanford's CoreNLP \cite{coreNLPtoolkit} successfully deliver the ability to automatically annotate unstructured texts with their underlying linguistic structures, including: Part-of-Speech (POS) Tags, Morphological Features, Dependency Relations, Named Entities, and so on. These annotations serve research labs, non-profit organizations and commercial  endeavors in their quest to {\em make sense} of the vast amount of unstructured data available to them. 

Universal processing pipelines such as UDPipe \cite{udpipe} aim to serve a range of other languages, but unfortunately, their performance on many morphologically rich languages (MRLs) \cite{tsarfaty10mrl}, and in particular Semitic languages, 
is not on a par with their performance on English. This, in turn, greatly limits their applicability for further research and commercial use. 
The main reason for this sub-optimal performance on Semitic languages is that the  {\em pipeline} design inherent in these frameworks is inappropriate for languages that exhibit extreme morphological ambiguity in their input stream. 
This is because errors made in morphological segmentation and disambiguation early on, jeopardize the system accuracy 
down the pipeline.
For Hebrew,
this performance gap has long been a {\em show-stopper} for advancing Language Technology and Artificial Intelligence for the Hebrew-speaking community. With this contribution, we aim to  remedy this situation.

In this paper we describe the design and use of the {\sc Onlp} system, 
a {\em joint} morphological-syntactic parsing framework for processing the Semitic language Modren Hebrew (Henceforth, Hebrew). The system is accurate, efficient, and provides rich and  expressive output including:  Segmentation, POS tags, Lemmas, Features and Labeled Dependencies. The {\em joint} training and  inference over the different  layers substantially limits error propagation, and leads in turn to speed and high accuracy. Among the technical advantages of the {\sc Onlp} suite are its open license, an easy 3-step installation, and a single package with all elements 
included  ---  no need to train or maintain individual components separately. 
The  {\sc Onlp} suite  already serves academic and commercial projects in diverse domains. Its accompanying online demo has further proved valuable for   educational purposes, exposing CS/NLP and non-CS researchers and engineers to the  intricacies of Semitic NLP.

\section{The Linguistic Challenge}
\label{sec:ling}
In morphologically-rich languages (MRLs), each  input token may consist of multiple lexical and functional units (henceforth, {\em morphemes}), each of which serves a particular role in the overall syntactic or semantic representation. In Hebrew, for example, the   token {\em `\cjRL{wk/smhm`bdh}`} corresponds to five  word tokens in English, each of which carrying its distinct  role:
{\em `\cjRL{w}`} (and, CC),
{\em `\cjRL{k/s}`} (when, REL),
{\em `\cjRL{m:}`} (from, IN), 
{\em `\cjRL{h}`} (the, DT),
{\em `\cjRL{m`bdh}`} (lab, NN).\footnote{We  use the annotation conventions of \citet{}{simaan01} that underlie the Hebrew SPMRL scheme \url{http://www.spmrl.org/spmrl2013-sharedtask.html}.}
This means that in order to process Hebrew texts, one first needs to segment  the Hebrew tokens into their constituting morphemes. At the same time, Hebrew raw tokens are highly ambiguous. A token such as: 
{\em `\cjRL{hqph}`}
may be interpreted as   
{\em `\cjRL{hqph}`} (orbit, NN),
{\em `\cjRL{h}` + `\cjRL{qph}`} (the+coffee, DT+NN), or  
{\em `\cjRL{hqp}'+ `\cjRL{/sl}' + `\cjRL{hy'}`} (perimeter of her, NN+POSS+PRP), etc. This is further complicated by  the lack of diacritics in standardized texts, meaning that most vowels are not present, and that no reading is  a-priory more likely than the others, out of context.  Only  {\em in context} the correct interpretation and segmentation become apparent. 

\begin{figure}
    \centering\scalebox{0.4}{
    \includegraphics{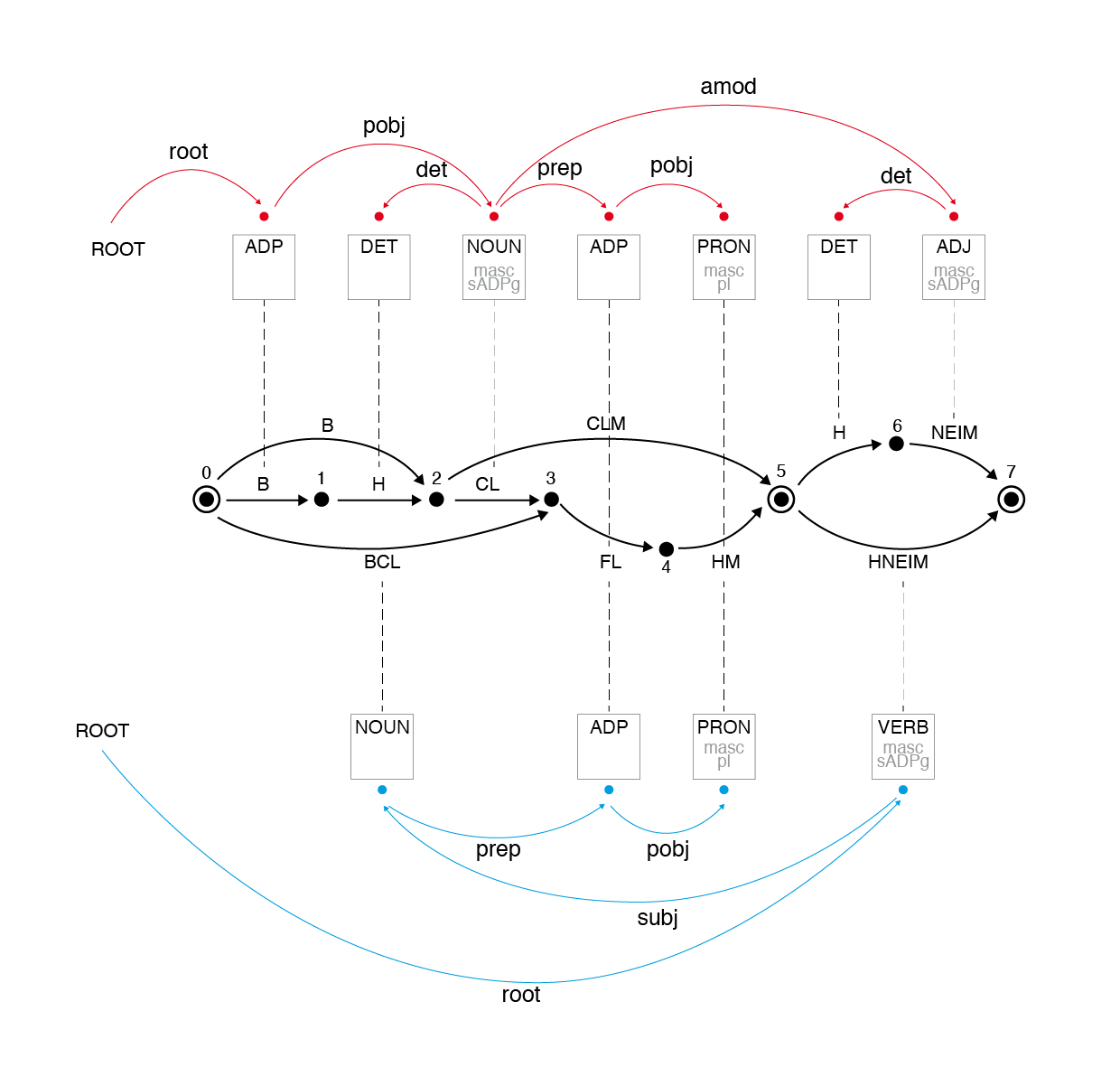}
    }\caption{The Joint Morpho-Syntactic  Search-Space. Lattice paths are of different lengths. Each lattice path can be assigned an exponential number of trees.}
    \label{fig:joint}
\end{figure}

These facts create an apparent loop in the design of NLP pipelines for Hebrew: {\em syntactic parsing  requires   morphological disambiguation  -- but   morphological disambiguation  requires   syntactic context}. This apparent loop 
has called for the development of {\em joint systems} rather than {\em pipelines}, for Semitic languages processing \cite{tsarfaty06,green10arabic}. This joint hypothesis has  proven useful for Hebrew and Arabic  phrase-structure parsing \cite{goldberg08joint,green10arabic,goldberg11}.
The {\sc Onlp} suite is a {\em dependency-based} parsing framework   implementing this joint hypothesis, over the entire morpho-syntactic search-space, as depicted in Figure \ref{fig:joint} \cite{more19}.

\section{The Architectural Design}
\label{sec:design}
The core of  {\sc Onlp}  is  {\em YAP (Yet Another Parser)}, a morpho-syntactic parser   for  morphological and syntactic analysis  of  Hebrew Texts. YAP re-implements and extends the  structure-prediction framework of \citet{zhang11framework}. We describe  {\em YAP}   in detail in \citet{more16coling,more19}. Here we  only provide  a bird's eye view  of the architecture.

In YAP we embrace the extreme morphological ambiguity in Hebrew. That is, we do {\em not} aim to resolve morphological ambiguity via pre-processing. The  input to YAP is  the complete {\em Morphological Analysis (MA)} of an input sentence \(x\), termed here MA\((x)\).
MA\((x)\) is a {\em lattice} structure, consisting of all possible morphological analysis possibilities of the input sentence, as seen in the middle of Figure \ref{fig:joint}. Each {\em arc} is a tuple specifying the {\em start-index}, {\em end-index}, the {\em form} of the  segment, its {\em part-of-speech, lemma, features}, and the {\em index of the raw token} the arc has originated from. An {\em arc} in the lattice can serve as a {\em node} in a syntactic dependency tree.  Each contiguous path in the lattice presents one valid morphological segmentation of the sentence, for which a dependency tree can be assigned, as in Figure \ref{fig:joint}. For each path in the lattice, there is an exponential number of dependency trees that are potentially applicable. 

We refer to the task of selecting the most likely lattice-path as {\em Morphological Disambiguation}  (MD), and  to the task of selecting the most likely dependency tree for a given path as {\em Dependency Parsing} (DEP). 
For an input sentence 
$x$, our goal is to {\em jointly} predict a single pair of MD$(x)$ and DEP$(x)$  that are consistent with one another, and form the most-likely  analysis of the sentence. 

The MD component is the transition-based {\em morphological parser} of \citet{more16coling}, which is formally based  on the   structure-prediction framework of \citet{zhang11framework}. MD accepts a sentence lattice MA(x) as input and delivers a selected sequence of arcs (morphemes) MD(x) as output.  The transition-based system for MD selects arcs for MD one at a time. It decodes the lattice using beam-search, and keeps the K-best paths at each step, scored according to morpheme-level and token-level features, weighted via structured-perceptron learning.

The  DEP component  is a re-implementation of the \citet{zhang11} dependency parser for English,   adapted for Hebrew. We assume an  Arc-Eager transition system and beam-search decoding. Feature weights  are learned via the structured perceptron. We employ a carefully-designed feature set that reflects linguistic   properties of Hebrew such as its rich morphological paradigms, flexible word-order,  agreement, etc.  This provides SOTA results on Hebrew dependency parsing, albeit in Oracle (i.e., gold morphology) scenario.

Seen that both the  MD and DEP realize the same formal framework and inherit from the same computational machinery,  we can easily {\em unify} them and treat the morpho-synactic task as a single objective. The transition systems are combined and the beam-search decoder interleaves  morphological and syntactic decisions.\footnote{For a complete formal exposition of the algorithm we refer the reader to \citet{more19}} Now morphological decisions may be affected by syntactic content, and vice versa.

\begin{figure}[t]
  \includegraphics[width=\linewidth]{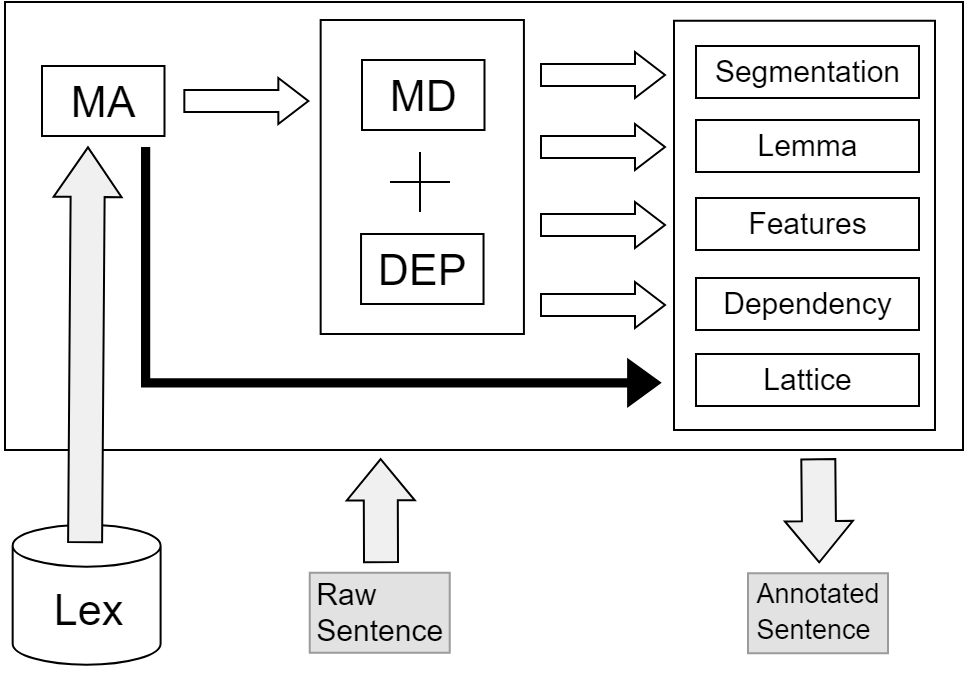}
  \caption{A bird's eye view of the Architecture}
  \label{fig:arch}
\end{figure}

The architecture is depicted in Figure \ref{fig:arch}.
In \citet{more19} we compared the performance of the joint system to our own pipeline system and to other    systems available for  Hebrew morphological and syntactic parsing, and showed significant improvements of YAP's joint model over all competing systems.

\section{The Annotation Scheme}
\label{sec:linlayers}

We deliver automatic morpho-syntactic  annotation of Hebrew texts based on the scheme of the SPMRL Hebrew dependency treebank.\footnote{The detailed annotation scheme is provided, with examples, in the supplementary material along with the screencast.}  The SPMRL Hebrew scheme employs the labels of \citet{simaan01} for  morphology and POS tags, and  the Unified-SD scheme of \citet{tsarfaty13} for the labeled dependencies.\footnote{With an eye for future comparability, we further developed a conversion algorithm to convert the the dependency tree from  Unified-SD to Universal Dependencies (UD).\url{https://universaldependencies.org/}}  
Specifically, 
we  deliver the following annotation layers:

\paragraph{Morphological Segmentation}
The most basic form of analysis of Hebrew texts is the segmentation of raw tokens into multiple meaning-bearing units that we call {\em morphemes}.
\footnote{In UD they are called {\em words}. In Hebrew NLP   they are called {\em segments}. We  use {\em morphemes} or {\em segments} herein.}

Due to  orthographic and phonological processes, some morphemes do not appear explicitly in the surface form. Our segmentation recovers all morphemes, both overt and covert.

the token \textit{`\cjRL{bbyt}'} (in the house) is  segmented as \textit{`\cjRL{b}' + `\cjRL{h}' + `\cjRL{byt}'}.

\paragraph{Part-of-Speech (POS) Tags}
Each morphological segment is assigned  a single Part-of-Speech tag category that indicates its syntactic role. 
The set of  tags used by the system is based on the SPMRL scheme which  in turn adopts the POS labels from \citet{simaan01} (detailed in our appendix).

\paragraph{Morphological Features} 
Along with the POS category, we specify  for each segment the properties that are signalled by inflectional morphology. The scheme encodes the following properties:
     {\bf Number} [S (Singular) / P (Plural) / D (Dual)], 
     {\bf Gender }[F (Female) / M (Male) / F,M (both)], 
     {\bf Person}  [1 / 2 / 3 / A (All)],\footnote{{\em A}  is used in cases where all analyses are valid, such as in Beinoni form - `\cjRL{'wklt}' (I/you/she eat.singular.feminine)} and
    {\bf Tense} [Past, Present, Future, Imperative, Infinitive].\footnote{Present-tense verbs and participles are tagged  `Beinoni'.}
    
\paragraph{Lemmas} 
Each segment is also assigned a lemma, i.e., the cannonical representation of its core (uninflected) meaning.\footnote{Note that due to high morphological fusion in Hebrew, simple surface-based  {\em stemming} will not suffice.} 
For Hebrew nouns and adjectives, the lemma is chosen to be the Masculine-Singular form. For verbs, the lemma is in the Masculine-Singular-3per form in Past tense.

\paragraph{Dependency Tree}
The dependency tree is defined over all morphological segments and an artificial root node. It consists of  a set of labeled binary relations that indicate the bi-lexical dependencies between segments.

Note that the SPMRL dependency scheme, as opposed to UD, always selects {\em functional} heads, rather than lexical heads. The  dependency labeling is based on the scheme from \citet{tsarfaty13}, repeated in the appendix. 

\paragraph{Lattices} 
As explained in section \ref{sec:design} above, a word can be segmented into morphemes in multiple ways, which are constrained by a broad-coverage lexicon. In addition to the parsed output, we makes available for each input sentence its sentence lattice, i.e. the set of all possible segmentations for a given sentence, along with all possible morphosyntactic analyses for each arc.
\section{Technical Details and Forms of Use}
\label{sec:usage}

YAP is implemented in the Go language.\footnote{\url{https://golang.org/}} It requires 6GB of RAM to run, and employs a simple 3-step installation, given in the supplementray material in the appendix.
The input to the system is a tokenized sentence, with tokens appearing one per line, and a line break after every sentence.\footnote{We assume the tokenization convention of MILA \cite{itai08}.} 
The output is a dependency tree (where each node in the tree is a lattice arc) provided in the CoNLL-X format \cite{conllX}.
 YAP is trained on the Hebrew section of the SPMRL  shared task. It also makes use of the broad-coverage lexicon of \citet{itai08} for finding all potential lattice paths. In case of out-of-vocabulary (OOV) items, we employ a simple heuristics where we suggest the 10 most-likely analyses of rare tokens observed during training.


\paragraph{Simple Use $|$ Command line}
From the command line, one can process one input file at a time, with a single sentence or more. 
The input file must be formatted with a single token per line, and an empty line denoting the end of every sentence.

Processing a file is done in 2 steps:
First, run Morphological Analysis \texttt{./yap hebma} to generates a sentence lattice containing all possible morphological breakdowns of each token. YAP will save the lattice to the file specified via  the \texttt{-out} flag.

Now you can run joint Morphological Disambiguation and Dependency Parsing \texttt { ./yap joint} to jointly predict the best lattice path and corresponding dependency tree. The input to this command is the output file generated in the previous step, and there are 3 output files: one containing word segments, one containing the  disambiguated lattice path,  and one containing the complete dependency tree in CoNLL-X format.

\paragraph{Advanced Use $|$ RESTful API}
\label{subsec:Ausage}
YAP can run as a RESTful server that accepts parse requests.  To do this simply start the server, listening on localhost port 8000.
Now you can call the joint endpoint with a json object containing the list of tokens to process in the HTTP data payload.
The response is a json object containing the three output levels (MA, MD and Dep). You can use jq and sed (or any other json and line processing tools) to format the (tab separated value) responses and reassemble the output. Check our appendix  for an illustration.

\paragraph{Educational Use $|$ The Online Demo}
\label{sec:onlineDemo}

In 2018 we decided to create an online demo of the system, for educational purposes:
    (i) To exposed NLP/AI researchers  to  NLP  capabilities available for  Hebrew.
    (ii) To educate non-CS scientists and engineers who work with Hebrew data (e.g., digital humanities)
    on text annotations that can potentially be useful for  their applications.
    (iii) To launch  outreach activities where we teach  {\em what is NLP} to the local community (e.g., school kids).\footnote{E.g., \url{https://www.youtube.com/watch?v=TFwQeoKpznA&feature=youtu.be}}
    
    To use the demo, simply go to \url{onlp.openu.ac.il} and type  Hebrew sentence in the textbox. 
The demo is built with Django and Bootstrap web frameworks. It sends the user's Hebrew text input to the {\sc Onlp} server, which returns a CoNLL-X formatted parse along with the  complete sentence lattice. 
Pre-processing includes pre-morphological tokenization of the input, where punctuation is being separated from the tokens. Double quotation marks are being separated from the word unless they appear before the last character of the word, to avoid over-segmentation of acronyms.\footnote{Acronyms in Hebrew are written with a quotation mark before the last letter, e.g. `\cjRL{b}"\cjRL{'rh}' (USA) .} The tokenized sequence is then passed to the {\sc Onlp} server.  
The CoNLL-X output is then processed into the following  layers: 
the FORM column is concatenated and presented as  "Segmented Text", and the POS, LEMMA, FEATS and DEPS are presented  in separate accordion tabs.

Furthermore, 
the demo  presents the sentence lattice which is the input to  the joint parser.
This is  useful for debugging, and for analyzing lexical-coverage in out-of-domain scenarios.

\paragraph{Expert Use $|$ Out of Domain Scenarios}
\label{sec:exlex}
A bottleneck for the system in out-of-domain parsing scenarios is the coverage of the lexicon. 
We rely on a general-purpose  lexicon containing over 500K entries. OOV words are treated via heuristics we designed, which are suitable for the general case only. 
However, identifying accurately  vocabulary items may be critical when applying the parser to new domains with domain-specific information (medical, financial, political, etc.). Fortunately, we can extend the system with a domain-specific lexicon, thus extending the MA coverage. Due to joint inference, the availability of a better suited {\em lexical} analysis triggers better {\em lexico-syntactic} decisions on the whole.\footnote{We discuss how exactly this is executed in the appendix.}

\section{Related and Future Work}

\begin{table}
\centering
\scalebox{0.65}{
\begin{tabular}{|l| ccccccc|c|}
\hline
\hline
  & Tok &MA   & MD & POS & Lem & Feats & Deps & Joint   \\
\hline
\hline
{\bf  Tasks} &\\

\hline
MILA &  $\checkmark$ &  $\checkmark$  & & &&& & \\
NITE &  $\checkmark$ &   $\checkmark$ &  $\checkmark$& & & &&\\
Hebrew-NLP & &  $\checkmark$ & && & &&  \\
Adler &  & &  $\checkmark$ & \checkmark& & \checkmark&& \\
Goldberg & &&&& $$&&\checkmark&\\
\hline
\hline
{\bf  Pipelines} &\\

\hline
UDPipe &  $\checkmark$ & $\checkmark$& $\checkmark$& $\checkmark$& $\checkmark$ & $\checkmark$ & $\checkmark$ &\\
CoreNLP &  $\checkmark$ & $\checkmark$& $\checkmark$& $\checkmark$& $\checkmark$ & $\checkmark$ & $\checkmark$ &\\\hline
ONLP   &  & $\checkmark$ & $\checkmark$& $\checkmark$& $\checkmark$& $\checkmark$& $\checkmark$ & $\checkmark$  \\ 
\hline
\hline
\end{tabular}}
\caption{Existing Coverage for Hebrew NLP Tasks}\label{tab:tasks}
\end{table}

Hebrew NLP in general and  Hebrew  parsing in particular are known to be challenging, due to  interesting linguistic properties, the scarcity of annotated data, and the small research community around. So, Hebrew has been seriously under-studied in NLP. During the early 2000, the MILA knowledge center was established, where the two of the main  Hebrew resources for NLP were developed: the Hebrew treebank \cite{simaan01} and the Hebrew Lexicon \cite{itai08}. 

Morphological Taggers for Hebrew using local linear-context have been trained on these data and were made available for free use \cite{adler06,barhaim06}. However, their performance was not on a par with parallel tools for English and thus  insufficient for commercial use. Hebrew dependency parsing was initially provided by \citet{goldberg09dep}, but the parser provides  {\em unlabeled} dependency, and the pipeline relied on Adler's morphological tagger. This left the automatic  dependency trees inaccurate and unsatisfying. 
{\em Joint} morpho-syntactic models for constituency-based parsing models  \citet{tsarfaty10rr} showed good performance on benchmark data, but their code was never released for open use.

With the development of the UD treebanks collection,   general frameworks such as UDPipe \cite{udpipe} and CoreNLP  \cite{coreNLPtoolkit} have been trained on the Hebrew UD treebank,  and made the model available. However, these  models provide performance that is still far from satisfactory, As we also demonstrate in our screen-cast,\footnote{\url{https://www.youtube.com/watch?v=H6pvh1x20FQ}} these systems make very basic mistakes, even with the simplest sentence. We conjecture that this is due to their inherent pipeline assumption: initial layers of processing present many mistakes. due to the extreme morphological ambiguity, and later layers cannot recover. Notably, also neural network models utilizing word embeddings, (e.g., UDPipe) still lag behind.

Table \ref{tab:tasks} shows the task-coverage of existing  tools and toolkits for NLP in Hebrew, academic as well as private initiatives (NITE,Hebrew-NLP). The task-coverage of the {\sc Onlp} suite we present is on a par with international standards (UDPipe, CoreNLP), and its level of performance was shown to exceed all existing models \cite{more19}. We are currently working towards  Named-Entity Recognition  as well as Open  Information Extraction, to be added to  {\sc Onlp}  in the near future.


\section{Conclusion}
\label{sec:conclusion}

This paper presents {\sc Onlp},  a complete language-processing framework for  automatic annotation of Modern Hebrew Texts. The framework covers morphological segmentation, POS tags, lemmas and  features, and dependency parsing, predicted jointly. The system is easy to install and to use, and we support multiple forms of usage fitting  user-personas with different needs. 
We hope the availability of an open-source, accurate, and easy-to-use system for NLP in Hebrew will benefit the local NLP open-source community and  greatly advance Hebrew language technology research and development, in academia and in the industry.

\section*{Acknowledgements}
We thank the NLPH community, in particular Shay Palachi, Amit Shkolnick and Yuval Feinstein, for much discussion and insightful comments. We further thank the Avi Bivas  (Innovation Authority) and Milo Avisar for promoting NLP initiatives in Israel. This research is supported by an ISF grant (1739/26) and an ERC  Starting grant (677352), for which we are grateful.

\bibliographystyle{acl_natbib}
\bibliography{main_and_supp}
\hfill \break \hfill \break

\appendix 
\section*{Supplementary Material For EMNLP Demo Paper}
These supplementary materials document {\em the absolute essentials} for starting to use the system: installation, annotation scheme documentation, forms of use, and enhancements for out-of-domains scenarios.

\footnotesize
\section{Resources}
\begin{itemize}
\item[] 1. YAP Github:\\ 
\url{https://github.com/OnlpLab/yap}\\
  2. YAP Demo - Website:\\
\url{http://onlp.openu.org.il}\\
  3. YAP Demo - Screencast: (Youtube)\\
\url{https://www.youtube.com/watch?v=H6pvh1x20FQ}
\\4. YAP Python-Wrapper:\\\url{https://github.com/amit-shkolnik/YAP-Wrapper
}
\\ 5.  SPMRL-to-UD Conversion: \\
\url{https://github.com/OnlpLab/Hebrew_UD}
\\ 6.  ONLP Lab Website: \\
\url{http://onlp.openu.org.il/home}
\end{itemize}

\section{Screen-Cast}
Check out our screen-cast online demo at:
\url{https://www.youtube.com/watch?v=H6pvh1x20FQ}
\section{Morphological Ambiguity: Lattices}
Table~\ref{lattice-example} shows a sentence lattice capturing the high  ambiguity of Hebrew morphological analysis. For a simple 3-tokens input sentence, 22 possible arcs present valid analyses of the various tokens. A single consecutive path through the lattice needs to be selected, for the sentence to be further processed by  syntactic parsers or   downstream  applications.

\section{Annotation Layers}
The annotation scheme provided by {\sc Onlp} corresponds to the Hebrew section of the SPMRL shared task. 2013-2014\footnote{\url{http://www.spmrl.org/spmrl2013-sharedtask.html}}
The Part-of-Speech Tags we employ are provided, along with illustrative examples, in Table~\ref{tab:postags}. The Dependency labels are defined and illustrated in Table~\ref{tab:deprel}.

\section{The Online Demo}
\begin{figure*}
\scalebox{0.5}{
  \begin{tabular}{cc}
  \includegraphics[width=\linewidth]{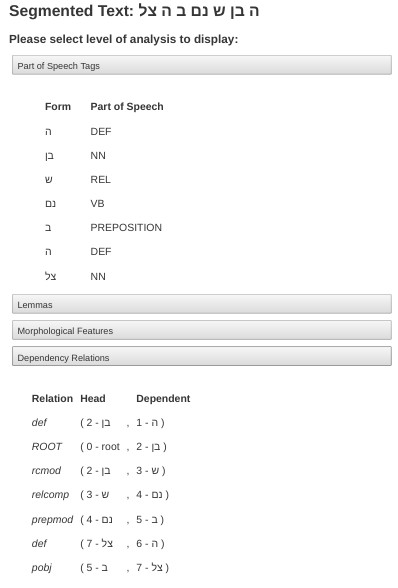}&
  \includegraphics[width=\linewidth]{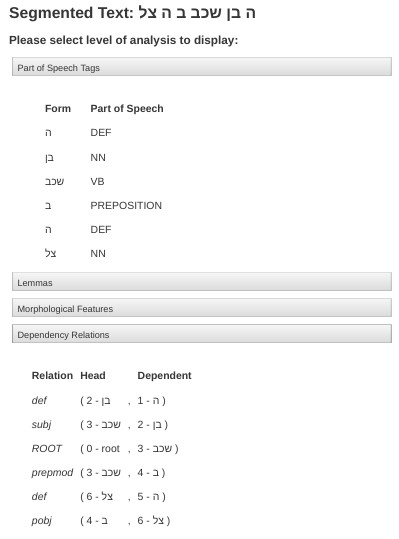} 
  \end{tabular}}
  \caption{On the right, we present a screen capture of the Morphological Segmentation, POS tags and Dependency Relations for the raw input sentence `\cjRL{hbn /skb b.sl}` ('the boy was lying in the shade'), as seen on our demo page. On the left,  we likewise present the Morphological Segmentation, POS tags and Dependency Relations for the nominal phrase `\cjRL{hbn /snm b.sl}` ('the boy that was napping in the shade'). Note that the two raw sentences have very similar form (in fact, they only differ in two characters). But they end up forming very different syntactic structures, which our system identifies and annotates correctly.
  }
  \label{fig:shaxav}
\end{figure*}

In Figure \ref{fig:shaxav}  we present a screen capture of the Morphological Segmentation, POS tags and Dependency Relations for two raw input sentences:
\begin{itemize}
    \item `\cjRL{hbn /skb b.sl}` \\'the-boy was-lying in-the-shade'
    \item  `\cjRL{hbn /snm b.sl}`\\ 'the-boy that-was-napping in-the-shade'
\end{itemize} As executed on our demo page.
Note that the two raw sentences have very similar form (in fact, they only differ in two characters). But they end up forming very different syntactic structures, which the {\sc Onlp} system  annotates correctly.
\section{Forms of Use}
Figures~\ref{fig:install}--\ref{fig:restful} present the usage patterns with the YAP parser, the core algorithm of the framework.
In Figure~\ref{fig:install} we present the 3-step installation, in  Figure~\ref{fig:cmd} we show a simple command-line use, and  in  Figure~\ref{fig:restful} we show how to use YAP as a service.
As noted before, The input file must be formatted with a single token per line and an empty line denoting end of sentence.\footnote{Crucially, the last line in the file {\em must} be empty to denote the end of the last sentence.}\footnote{A note for Windows users:  
YAP doesn't handle Windows style text files that have BOM marks and CRLF newlines. So if you're running on Windows and YAP doesn't work, make sure you don't have CRLF line endings and no BOM marks.}  YAP has been written in Go in order to enable multi-threading. This means that it can be called from multiple  threads in parallel. As of June 2019 there is also a python wrapper, created by members of the Israeli open-source community.\footnote{The Credit goes to Amit Shkolnik of the 4girls initiative. Further details can be found here: \url{https://github.com/amit-shkolnik/YAP-Wrapper}}
\section{Out-of-Domain Scenarios}

When observing errors in a new domain, one first thing we have to check is whether or not these are due to {\em lexical gaps}. I.e., whether they stem from lack of coverage of the lexicon. The availability of the sentence lattice output  is of great value in this respect.
By reviewing the lattice, it is possible to see whether  the lexicon contains the correct morphological analysis for the input token at all. If the correct analysis is not in the lattice, it  is easy to add the missing analyses by editing the lexicon.\footnote{The lexicon file located at \url{ data/bgulex/bgulex.utf8.hr}}

Each line in the lexicon file contains a token followed by a list of one or more possible morphological analyses of that token.
 An analysis is a tuple made of 3 parts $\langle$ {\em prefix:host:suffix} $\rangle$ followed by the {\em host} lemma.
Each tuple member contains the part-of-speech tag and morphological features for any of these elements. {\em prefix} and {\em suffix} can possibly be empty.
E.g.\
\\ 
$>$ {\em \cjRL{''bd} :VB-MF-S-1-FUTURE-NIFAL: \cjRL{n'bd} :VB-MF-S-1-FUTURE-PIEL: \cjRL{'ybd}}
 
An example use case could arise when processing medical domain texts related to cancer in which the word  {\em `\cjRL{lymph}` (lymph)} appears in the text but is missing from the lexicon. In this case, the parser errs in identifying the first `\cjRL{l}' as the preposition "to", followed by a proper noun. 

To remedy this, we can   update the lexicon by adding the following line:
 \\
$>$ {\em \cjRL{lymph} :NN-F-S: \cjRL{lymph} }\\
This means that the token \cjRL{lymph} is a common noun with feminine gender and singular number, followed by the lemma, and that it is unambigous (i.e., only one analysis is available). Note that after updating the lexicon you need to restart YAP (if running as a restful server) for the lexical changes to apply.

Now that \cjRL{lymph} is no longer an OOV, sentences containing this token will be given a more accurate lattice and as a result will be analyzed with a global syntactic structure that accords with the correct analysis. We suggested these lexicon edits for our  users working in specific domains in the industry (medical, social, political), and they attested to significant improvements when running on  particular domains.\footnote{Yuval Feinstain,  NLP Consultant, p.c.} 

\begin{table*}
\centering
\scalebox{0.85}{
\begin{tabular}{c c c c c c c }
\hline
\hline
From &To   & Form & Lemma & Part of Speech & Features & Token Number   \\
\hline
\hline
0 & 1 & \cjRL{h} & \cjRL{h} & DEF & \_ & 1  \\
0 & 3 & \cjRL{h} & \cjRL{h} & REL & \_ & 1  \\
0 & 5 & \cjRL{hbn} & \cjRL{hbyn} & VB & gen=M,num=S,per=2,tense=IMPERATIVE & 1  \\
1 & 2 & \cjRL{b} & \cjRL{b} & IN & \_ & 1  \\
1 & 5 & \cjRL{bn} & \cjRL{bn} & NNP & gen=M,num=S & 1  \\
1 & 5 & \cjRL{bn} & \cjRL{bn} & NNT & gen=M,num=S & 1  \\
1 & 5 & \cjRL{bn} & \cjRL{bn} & NN & gen=M,num=S & 1  \\
2 & 5 &  \cjRL{hn} & \cjRL{hn} & S\_PRN & gen=F,num=P,per=3 & 1  \\
3 & 4 &  \cjRL{b} &  \cjRL{b} & IN & \_ & 1  \\
3 & 5 & \cjRL{bn} & \cjRL{bn} & NNP & gen=M,num=S & 1  \\
3 & 5 & \cjRL{bn} & \cjRL{bn} & NNT & gen=M,num=S & 1  \\
3 & 5 & \cjRL{bn} & \cjRL{bn} & NN & gen=M,num=S & 1  \\
4 & 5 & \cjRL{hn} & \cjRL{hn} & S\_PRN & gen=F,num=P,per=3 & 1  \\
5 & 6 & \cjRL{/s}  & \cjRL{/s} & REL & \_ & 2  \\
5 & 7 & \cjRL{/snm}  & \cjRL{/sn}  & NN & gen=F,num=S,suf\_gen=M,suf\_num=P,suf\_per=3 & 2  \\
6 & 7 & \cjRL{nm} & \cjRL{nm} & VB & gen=M,num=S,per=A,tense=BEINONI & 2  \\
6 & 7 & \cjRL{nm} & \cjRL{nm} & BNT & gen=M,num=S,per=A & 2  \\
6 & 7 & \cjRL{nm} & \cjRL{nm} & BN & gen=M,num=S,per=A & 2  \\
6 & 7 & \cjRL{nm} & \cjRL{nm} & VB & gen=M,num=S,per=3,tense=PAST & 2  \\
7 & 8 &  \cjRL{b} &  \cjRL{b} & PREPOSITION & \_ & 3  \\
7 & 10 &  \cjRL{b.sl} & \cjRL{b.sl} & NN & gen=M,num=S & 3  \\
7 & 10 &  \cjRL{b.sl} &  \cjRL{b.sl} & NNT & gen=M,num=S & 3  \\
8 & 9 & \cjRL{h} & \cjRL{h} & DEF & \_ & 3  \\
8 & 10 & \cjRL{.sl}  &  \cjRL{.sl} & NN & gen=M,num=S & 3  \\
8 & 10 & \cjRL{.sl}  & \cjRL{.sl}  & NNT & gen=M,num=S & 3  \\
9 & 10 & \cjRL{.sl}  & \cjRL{.sl}  & NNT & gen=M,num=S & 3  \\
9 & 10 & \cjRL{.sl}  & \cjRL{.sl}  & NN & gen=M,num=S & 3  \\
\hline
\hline
\end{tabular}
}
\caption{The 
Lattice representation for `\cjRL{hbn /snm b.sl}` `The boy who slept in the shade'.
    Col 1-2: From/To - the start and end nodes of the morpheme. The numbers are with respect to the maximal length route.
    Col 3: Form - the surface form of the morphological segment.
   Col 4-5-6: Form/Lemma/Part of Speech - the same segment may belong to different entries in the lexicon. Each entry is given in a separate row, where the differences between the different meanings are surfaced in one (or more) of the Form/Lemma/Part of Speech columns. 
    Col 7: Token Number - represents the index of the raw (space-delimited) token in the input before segmentation.
}
\label{lattice-example}

\end{table*}
\centering
\begin{table*}
\scalebox{0.75}{
\begin{tabular}{l  l  r}
\hline
POS & Definition & Example \\
\hline
\hline
ADVERB & The word \cjRL{k*:} before numerals &  \cjRL{k*:mylywn} \\
AT & The accusative marker \cjRL{'t} which is a seperate word in Hebrew &  \cjRL{'t hklb} \\
BN & Participle (Beinoni) &  \cjRL{mgy`ym} \\
BNT & Participle in construct state form &  \cjRL{mqymy h`ytwn} \\
CC & Conjunction &  \cjRL{'l'} \\
REL & Relative clause marker &  \cjRL{/s:} \\
CD & Numeral &  \cjRL{m'wt} \\
CDT & Numeral in construct state &  \cjRL{'lpy} \\
CONJ & Coordinating conjunction \cjRL{w} &  \cjRL{w:} \\
COP & Copula &  \cjRL{hyh} \\
DEF & A special tag assigned to the definite marker \cjRL{h} which appears with nouns, adjectives and numerals &  \cjRL{h} \\
DTT & Determiner &  \cjRL{kl} \\
DUMMY\_AT & Accusative marker \cjRL{'t} when used with a pronominal suffix &  \cjRL{'wtw} \\
EX & The existential markers \cjRL{y/s} or \cjRL{'yn} &  \cjRL{y/s} \\
IN & Preposition &  \cjRL{`d} \\
INTJ & Interjection &  \cjRL{n'} \\
JJ & Adjective &  \cjRL{zrym} \\
JJT & Adjective in construct state &  \cjRL{ypy np/s} \\
MD & Modal predicates &  \cjRL{.sryK} \\
NN & Noun &  \cjRL{.hbr} \\
NN\_S\_PP & Noun with a pronominal suffix &  \cjRL{pw`lyhM} \\
NNP & Proper Noun &  \cjRL{n`my} \\
NNT & Construct state noun &  \cjRL{h`sqt} \\
P & Prefix written as a separate word &  \cjRL{blty} \\
POS & Possessive preposition \cjRL{/sl} &  \cjRL{/sl} \\
PREPOSITION & Inseperable preposition &  \cjRL{b*:} \\
PRP & Personal Pronoun &  \cjRL{hy'} \\
S\_PRP & Reflexive pronoun &  \cjRL{`.smy} \\
QW & Question word &  \cjRL{ky.sd} \\
S\_PRN & Personal pronoun attached to a preposition as a pronominal suffix &  \cjRL{'wtnw} \\
TEMP & Subordinating conjunction introducing time clauses &  \cjRL{k*:/s:} \\
VB & Verb &  \cjRL{'mrh} \\
yyCLN & Colon &  : \\
yyCM & Comma &  , \\
yyDASH & hyphen or dash &  - \\
yyDOT & Period &  . \\
yyELPS & Ellipsis &  ... \\
yyEXCL & Exclamation mark &  ! \\
yyLRB & Left Parenthesis & \big( \\
yyQM & Question Mark &  ? \\
yyQUOT & Quotation Mark & " " \\
yyRRB & Right Parenthesis & \big) \\
yySCLN & Semicolon &  ; \\
\hline \hline
\end{tabular}
}
\caption{The Part-of-Speech Tags Provided by {\sc Onlp}}\label{tab:postags}
\end{table*}

\begin{table*}
\centering
\scalebox{0.75}{
\begin{tabular}{l l l r}
\hline
Dependency & Definition & Example \\
\hline
\hline
num & numerical modifier & num (\cjRL{'n/syM}, \cjRL{`/srwt})  & \cjRL{`/srwt 'n/syM mgy`ym mt'ylnd}\\
subj & subject & subj (\cjRL{htbrrh}, \cjRL{htwp`h})  & \cjRL{htwp`h htbrrh 'tmwl} \\
ROOT & root & ROOT ( root ,\cjRL{.t`nh})  & \cjRL{hy' .t`nh kK}\\
prepmod & prepositional modifier & prepmod (\cjRL{m:} , \cjRL{.sd}) & \cjRL{m.sd '.hd}\\
pobj & object of a preposition & pobj (\cjRL{l:} , \cjRL{n.sygyM}) & \cjRL{hy' tpnh ln.sygym}\\
comp &  complement & comp (\cjRL{bkh}, \cjRL{k'/sr}) & \cjRL{hyld bkh k'/sr lq.hw lw 't h.s`.sw`}\\
conj & conjunct & conj (\cjRL{w:} , \cjRL{r`myM} ) & \cjRL{r`myM wbrqyM}\\
punct & punctuation & punct (\cjRL{n/sm`h} , : ) & !\cjRL{ttkwpP} :\cjRL{n/sm`h qry'h} \\
advcl & adverbial clause & advcl (\cjRL{'M}, \cjRL{yw/sg}) & \cjRL{hM y/sbtw 'M l' yw/sg hskM}\\
advmod & adverbial modifier & advmod (\cjRL{ytqblw}, \cjRL{l'ltr}) & \cjRL{kwlM ytqblw l'ltr}\\
obj & object & obj (\cjRL{t`/sh}, \cjRL{ml.hmh}) & \cjRL{bc.hkmh t`/sh lk ml.hmh}\\
amod & adjectival modifier & amod (\cjRL{hby.tw.h},\cjRL{hl'wmy}) & \cjRL{hby.tw.h hl'wmy b/sbyth}\\
det & determiner & det (\cjRL{hyldyM},\cjRL{kl}) & \cjRL{'ny rw'h 't kl hyldyM}\\
def & definite marker & def (\cjRL{'mbwlns},\cjRL{h}) & \cjRL{h'mbwlns ht.hyl lnsw`}\\
gobj & genitive object & gobj (\cjRL{pr/sy},\cjRL{m/s.trh}) & \cjRL{pr/sy m/s.trh y.s'w}\\
possmod & possession modifier & possmod (\cjRL{w`dt},\cjRL{/sl}) & \cjRL{w`dt hkspyM /sl hknst}\\
rcmod & relative clause modifier & rcmod (\cjRL{hw`dh},\cjRL{/s:}) & \cjRL{hw`dh /sdnh bnw/s'}\\
relcomp & relative complement & relcomp (\cjRL{/s:},\cjRL{dnh}) & \cjRL{hw`dh /sdnh bnw/s'}\\
appos & apposition / parenthetical & appos (\cjRL{k*}"\cjRL{.h},\cjRL{mpM}) & (\cjRL{mpM} ) \cjRL{y'yr .sbN} \cjRL{k*}"\cjRL{.h} \\
nn & noun modifier & nn (\cjRL{sN},\cjRL{symwN}) & \cjRL{mnzr sN symwN}\\
ccomp & complement clause with internal subject & ccomp (\cjRL{/s:},\cjRL{.hmwd}) & \cjRL{'mrty lk /s'th .hmwd}\\
neg & negative modifier & neg (\cjRL{tk`s},\cjRL{l'}) & \cjRL{hy' l' tk`s}\\
pcomp & complement clause of a preposition & pcomp (\cjRL{kdy},\cjRL{lh/sttP}) & \cjRL{hw' .ts kdy lh/sttP bt.hrwt}\\
xcomp & complement clause with external subject & xcomp (\cjRL{r.sh},\cjRL{lh`lwt}) & \cjRL{hw' r.sh lh`lwt 't h/skr} \\
acc & accusative case & acc (\cjRL{ly.tpty},\cjRL{'t}) & \cjRL{ly.tpty 't hklb}\\
vmod & verb as modifier & vmod (\cjRL{sykwy},\cjRL{lhtqbl}) & \cjRL{y/s lw sykwy lhtqbl l'qdmyh}\\
gen & genitive case & gen (\cjRL{mktbh},\cjRL{/sl}) & \cjRL{mktbh /sl mly pylypsbwrN}\\
number & numerical modifier in digits & number(\cjRL{.htymwt},84) & \cjRL{.htymwt} 84 \cjRL{hw' 'sP}\\
mwe & multi-word expression & mwe (\cjRL{mdy},\cjRL{/snh}) & \cjRL{hw' .ts mdy /snh}\\
goeswith & tokens originally connected with a hyphen & goeswith (\cjRL{mwnswn},\cjRL{nwwh}) & \cjRL{mwnswn}-\cjRL{'n.hnw gryM bnwwh}\\
cop & copular element & cop (\cjRL{mqwM},\cjRL{hy'}) & \cjRL{'ywbh hy' mqwM l' /sgrty}\\
cc & introducing conjunction & cc (\cjRL{'mr},\cjRL{hry}) & \cjRL{hry hw' 'mr z't qwdM}\\
npred & noun as predicate & npred (\cjRL{hyh},\cjRL{qwmwnys.t}) & \cjRL{hw' hyh qwmwnys.t}\\
parataxis & side-by-side, interjection & parataxis (\cjRL{'/sM},\cjRL{nwld}) & \cjRL{hw' nwld kkh} ,\cjRL{hw' l' '/sM} \\
npadvmod & noun phrase as adverbial modifier & npadvmod (\cjRL{yhyh},\cjRL{ywM}) & \cjRL{ywM '.hd hw' yhyh hn/sy'}\\
apred & adjective as predicate & apred (\cjRL{hyyty},\cjRL{tmyM}) & \cjRL{mstbr /shyyty tmyM}\\
vocative & explicitly addressing a dialogue participant & vocative(\cjRL{h`t},\cjRL{rbwty}) & \cjRL{rbwty} ,\cjRL{zw h`t ly/swN}\\
aux & auxilary verb or feature-bundle & aux (\cjRL{/sqw`h},\cjRL{hyth}) & \cjRL{klklth hyyth /sqw`h bmytwn}\\
ppred & preposition as predicate & ppred (\cjRL{yhyh},\cjRL{b*:}) & \cjRL{m.hr hq.tyP yhyh b`y.swmw}\\
acomp & adjectival complement & acomp (\cjRL{nr'h},\cjRL{myw.hd}) & \cjRL{hw' nr'h myw.hd}\\
qmark & question & qmark (\cjRL{ykwlyM},\cjRL{h'M}) & \cjRL{h'M 'tM ykwlyM lhm/syk}\\
\hline\hline
\end{tabular}
}\caption{The Dependency Labels Provided by {\sc Onlp}}\label{tab:deprel}
\end{table*}

\begin{figure}[t]
\begin{itemize}
\item[]
1. Setup a Go environment:
\\
\texttt{ mkdir yapproj ; export GOPATH=/full/path/to/yapproj; cd yapproj }
\\
2. Clone the repository in src folder:\\
\texttt{mkdir src; cd src; git clone github.com/OnlpLab/yap.git }
\\
3. Unzip the models and build yap: \\
\texttt{cd yap; bunzip2 data/*.bz2; go get .; go build.}
\end{itemize}
\caption{A  3-Step Installation. To install YAP make sure you have Go, Git and BZip2 installed and available on your system's PATH. The instructions are for Linux but similarly  can be done on Windows/MacOS}
\label{fig:install}
\end{figure}

\begin{figure}[t]
\centering
 \begin{itemize}
 \item[] 1. Morphological Analysis:\\
\texttt{ ./yap hebma -raw input.txt -out input.lattice }
\\ 2. Joint Morpho-syntactic Parsing: \\
\texttt { ./yap joint -in input.lattice -os output.segmentation -om output.mapping -oc output.conll }
 \end{itemize}
 \caption{Simple Use $|$ Command line}\label{fig:cmd}
\end{figure}

\begin{figure}[t]
    \centering
    \begin{itemize}
        \item[] 1. Start the server:\\
        \texttt{./yap api }\\
        2. Call the joint endpoint:\\
        \texttt{curl -s -X GET -H 'Content-Type: application/json' -d'{"text": "<word1> <word2> <word3> <word4> ..."}' localhost:8000/yap/heb/joint > response.json }
\\
3. The response is a jason object:\\
\texttt {jq '.ma\_lattice, .md\_lattice, .dep\_tree' < response.json | sed -e 's/\textasciicircum.//' -e 's/.\$//' -e 's/\textbackslash \textbackslash t/\textbackslash t/g' -e 's/\textbackslash \textbackslash n/\textbackslash n/g'}
    \end{itemize}
    \caption{Advanced Use $|$ RESTful API}
    \label{fig:restful}
\end{figure}

\begin{table*}
\centering
\scalebox{0.75}{
\begin{tabular}{l l l l}
\hline
& Morphological Analysis Lattice (.ma and .md files) & & \\
\hline
\hline
Column & Definition & Tag & Comment \\
\hline
col 1 & Morpheme Start Index in the Lattice & FROM & \\
col 2 & Morpheme end Index in the Lattice & TO & \\
col 3 & Form of the Morpheme & FORM & \\
col 4 & Lemma of the Morpheme & LEMMA & \\
col 5 & Coarse Part of Speech Tag & CPOSTAG & underscore if unavailable\\
col 6 &  Fine Part of Speech Tag & POSTAG & CPOSTAG and POSTAG are identical in YAP\\
col 7 & Morphological Features & FEATS & underscore if unavailable\\
col 8 &  Source Token Index & TOKEN & \\
\hline
& CONLL File format (.conll) & & \\
\hline
\hline
col 1 & Morpheme Index in the Sentence & ID & \\
col 2 & Form of the Morpheme & FORM & \\
col 3 & Lemma of the Morpheme & LEMMA & underscore if unavailable\\
col 4 & Coarse Part of Speech Tag & CPOSTAG & underscore if unavailable\\
col 5 &  Fine Part of Speech Tag & POSTAG & CPOSTAG and POSTAG are identical in YAP\\
col 6 &  Morphological Features & FEATS & underscore if unavailable\\
col 7 & Head Index Pointer & HEAD & note that the resulting structure is a tree \\
col 8 & Dependency relation to the HEAD & DEPREL & \\
col 9 & Projective Head & PHEAD & ignore - unused by YAP \\
col 10 & Dependency relation to the PHEAD & PDEPREL & ignore - unused by YAP \\
\hline\hline
\end{tabular}
}\caption{Columns Definitions in .ma, .md and .conll files}\label{tab:deprel}
\end{table*}

\end{document}